\documentclass[cameraready]{Interspeech}

\title{OpenWER: Improving Cross-Lingual ASR Evaluation and Enabling Token-Based Accuracy Metrics}

\author[affiliation={1,2}, orcid=0009-0005-1296-4987]{Korbinian}{Kuhn}
\author[affiliation={1}, orcid=0000-0002-3129-1897]{Gottfried}{Zimmermann}
\email{kuhnko@hdm-stuttgart.de, gzimmermann@acm.org}
\address{
    $^1$ Stuttgart Media University, Germany \\
    $^2$ University of Tübingen, Germany
}
\email{korbiniankuhn@gmail.com, gzimmermann@acm.org}


\usepackage{booktabs}
\usepackage[normalem]{ulem}
\usepackage [english]{babel} 
\usepackage{acronym}
\newacro{ASR}{Automatic Speech Recognition}
\newacro{WER}{Word Error Rate}
\newacro{CER}{Character Error Rate}
\newacro{WIL}{Word Information Lost}
\newacro{MER}{Match Error Rate}
\newacro{DHH}{d/Deaf and Hard of Hearing}
\newacro{E2E}{end-to-end}
\newacro{NLP}{Natural Language Processing}
\newacro{POS}{Part-of-Speech}
\newacro{NER}{Named-Entity Recognition}
\newacro{ML}{Machine Learning}
\newacro{JIT}{Just-in-Time}

\begin{document}

\maketitle

\begin{abstract}
    Advances in deep learning and end-to-end Automatic Speech Recognition (ASR) have enabled robust multilingual models, but evaluation metrics remain limited in assessing accuracy. Efforts to improve or replace the common metric Word Error Rate (WER) often focus on English, leaving evaluations for low-resource languages under-explored and hindering fair cross-lingual comparisons. We present OpenWER, an open-source implementation that improves WER robustness through language-specific normalisation and compound word detection. A token-based Levenshtein alignment preserves complementary metrics and allows metadata embedding for granular accuracy scores. Our analysis of 52 languages shows absolute WER reductions of up to 25\% compared to common libraries. OpenWER contributes to fairness in ASR research by increasing the reliability of WER across diverse languages and enabling more comprehensive accuracy evaluations.
\end{abstract}

\section{Introduction}

Large-scale deep learning techniques \cite{Rasmus2015, Baevski2020, Yu2020A} and transformer-based \ac{ASR} \cite{Vaswani2017, Chorowski2015, Gulati2020, Peng2022} have led to robust multilingual models that support a wide range of languages \cite{Radford2023, Barrault2023}. Evaluating their transcription accuracy using universal metrics is essential for unbiased and fair performance comparisons across models and languages. The predominant metrics for \ac{ASR} evaluation are \ac{WER} for alphabetic languages, where words consist of individual letters with clear boundaries, and \ac{CER} for logographic languages, where symbols represent words or morphemes. Both metrics rely on the Levenshtein distance, which quantifies differences between a hypothesis and a reference transcript at the word or character level \cite{Levenshtein1966}. Their simplicity has led to widespread adoption and enables automated computation on large datasets while offering an objective, language-independent assessment of transcription accuracy.

\ac{WER} has been criticised as an accuracy measure because of its weak correlation with human-perceived transcription quality \cite{Wang2003, Mishra2011, Favre2013}. As a purely string-based metric, it lacks linguistic and semantic knowledge, treating all errors equally, regardless of their impact on comprehension \cite{Berke2018}. \Ac{ML} addresses these limitations by classifying word importance to determine the severity of errors \cite{Kafle2016, Amin2023}. \Ac{NLP} techniques such as \ac{POS} tagging and \ac{NER} allow the categorisation of words, while statistical models can weight different error types \cite{Roux2022, Apone2010}. Cosine similarity of word embeddings allows semantic comparison of word substitutions \cite{Roux2022} and can be combined with word predictability derived from n-gram models or entropy scores \cite{Kafle2017, Kafle2019}. While \ac{ML}-based metrics offer potential improvements over \ac{WER}, they struggle with out-of-domain data and do not necessarily perform better than \ac{WER} \cite{Wells2022, Chavez2024}. Reliance on training data can introduce bias into evaluations and hinder adoption in low-resource languages, raising concerns about fairness in \ac{ASR} research.

Improvements to \ac{WER} have also been explored. \ac{MER} and \ac{WIL} quantify Levenshtein operations differently to overcome theoretical limitations of \ac{WER}, but produce similar scores for state-of-the-art \ac{ASR} \cite{Morris2004}. Text pre-processing is an effective approach to reducing non-semantic differences. Common strategies include removing punctuation and converting all characters to lowercase. Language-specific normalisation reduces variability by replacing contractions and spelling differences \cite{Koenecke2020}. Unifying written numbers can further reduce \ac{WER}, but remains challenging and can introduce errors \cite{Radford2023}. As language-specific normalisations are mainly available for English, they complicate cross-lingual comparisons. Moreover, the destructive nature of text pre-processing eliminates certain aspects of accuracy. Even with a low \ac{WER}, \ac{ASR}-generated text can be difficult to read due to punctuation, casing, and formatting errors \cite{Butler2019}. Token-based alignment allows the calculation of \ac{WER} and complementary metrics for punctuation and casing, providing a more comprehensive assessment of accuracy \cite{Meister2023, Kuhn2024}. Extending the Levenshtein distance algorithm to handle transpositions or compound words can also reduce alignment errors \cite{Damerau1964, Kuhn2024}.

While \ac{WER} is a limited measure of accuracy \cite{Wang2003, Mishra2011, Favre2013}, it remains the most common evaluation metric, as no alternative has yet been widely adopted \cite{Apone2010, Kafle2019, Roux2022}. Reducing noise from string differences is key to improving \ac{WER} robustness and enabling more meaningful scores \cite{Koenecke2020, Radford2023}. This study presents OpenWER, an open-source library that improves accuracy evaluation across high- and low-resource languages. The robustness of the Levenshtein distance is improved by non-destructive normalisation and compound word detection, reducing minor variations that distort \ac{WER} scores \cite{Kuhn2024}. The token-based alignment allows the calculation of complementary accuracy metrics and the integration of metadata from \ac{NLP} parsers or \ac{ASR} models, supporting more granular and comprehensive evaluations \cite{Roux2022, Meister2023}. Our results show:

\begin{itemize}
    \item OpenWER can decrease WER across languages up to 25\% compared to commonly used libraries.
    \item Compound word errors can reduce WER by up to 20\% for some languages.
    \item Token-based alignment preserves punctuation, casing, and embedded metadata for complementary metrics.
\end{itemize}

\begin{figure*}[ht!]
  \centering
  \includegraphics[width=\linewidth]{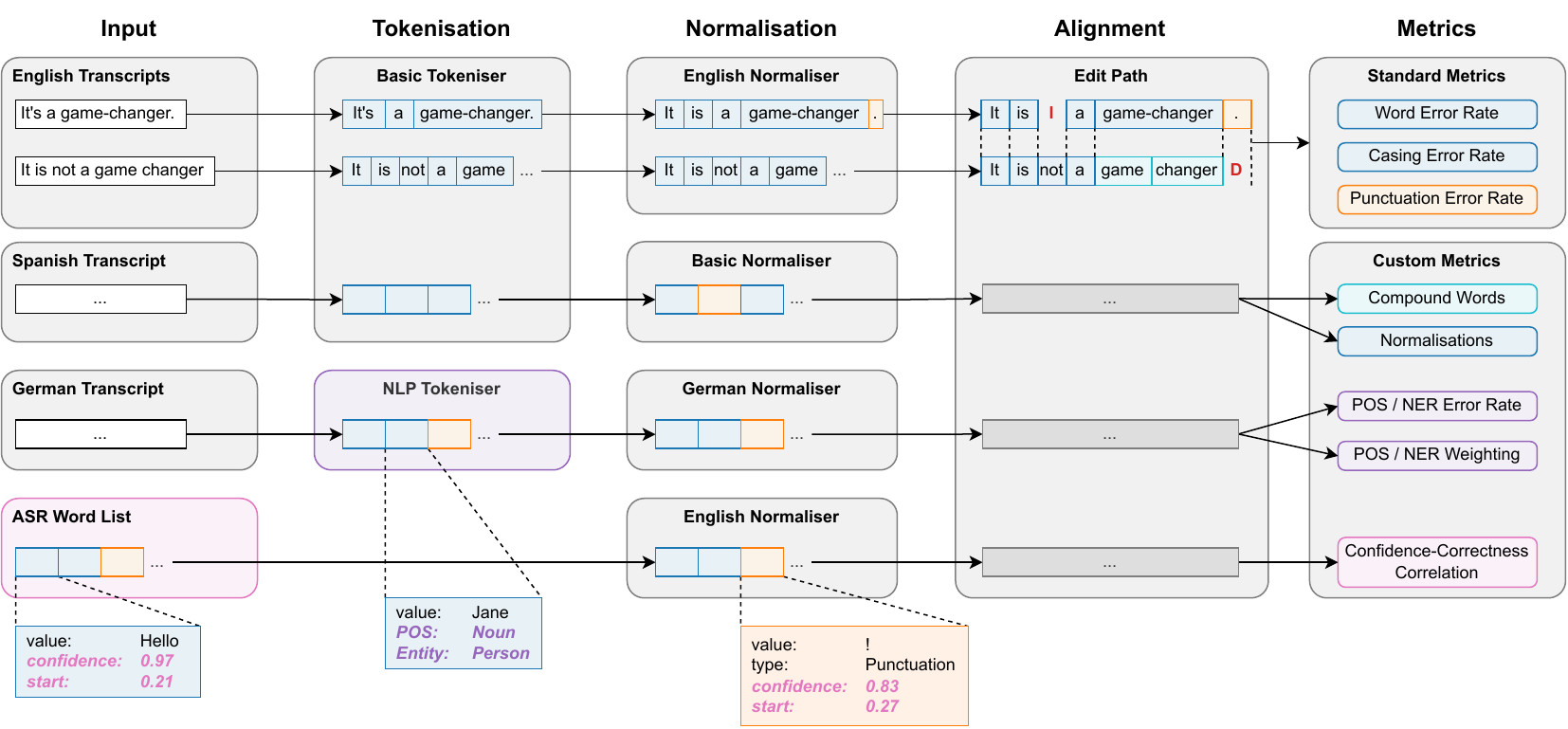}
  \caption{\textbf{Overview of OpenWER components}: Input can be raw text or ASR-generated word lists. Basic tokenisation is built in, but can be replaced by an NLP-tokeniser. General or language-specific normalisation can be applied to all input and tokenisation methods. Hypothesis and reference are aligned using an extended Levenshtein distance algorithm that handles punctuation tokens and compound words. After alignment, the Word Error Rate and complementary metrics for punctuation and casing can be computed. More granular token-based metrics can be derived from ASR or NLP metadata.}
  \label{fig:approach}
\end{figure*}

\section{OpenWER}

This section outlines the components of OpenWER. Figure \ref{fig:approach} illustrates its modular design, with interchangeable elements for input formats, tokenisation methods, language-specific normalisation, and evaluation metrics.

\subsection{Input}

The computation of the edit distance requires a reference and hypothesis input. These inputs can be raw text or a structured list of words. Formats may also differ to enable scenarios where speech datasets provide the reference as raw text and \ac{ASR} models provide the hypothesis as a word list. Additional properties, such as timestamps or confidence scores in the word list, are preserved throughout the computation, allowing for custom evaluations after alignment.

\subsection{Tokenisation}

Unlike traditional string list comparisons used to compute the Levenshtein distance, OpenWER uses tokens as alignment units that preserve the original text and metadata. This approach allows embedding additional information throughout the computation process and quick differentiation between words and punctuation characters. A tokeniser segments text at spaces, tabs and newlines. Punctuation remains attached during tokenisation and is handled later in a normalisation step, allowing for preceding transformations such as abbreviation handling. \ac{NLP} tokenisers can be used to include \ac{POS} tags or \ac{NER} labels as metadata, allowing fine-grained evaluation metrics.

\subsection{Normalisation}

Text pre-processing can be applied to all input methods, including basic or \ac{NLP}-based tokenisation and \ac{ASR} word lists. While normalisation is optional, reducing non-semantic differences between reference and hypothesis is recommended to improve the Levenshtein alignment's robustness and ensure more meaningful error rates. Normalisation consists of sequential transformations applied to token values, mostly affecting individual tokens, although some may split or merge tokens. Metadata and original token values are preserved throughout the process, and all modifications are tracked in a token's normalisation field. The order of transformations is important; for example, abbreviations must be normalised before separating punctuation. A default normaliser with general transformations and language-aware punctuation separation is provided. Language-specific normalisers based on related research are available for English and German \cite{Koenecke2020, Radford2023}.

\subsection{Alignment}

The alignment of reference and hypothesis tokens is performed using the Levenshtein distance. We adopt a variable cost approach inspired by previous work to improve the semantic relevance of edit operations \cite{Kuhn2024}. Punctuation edits receive lower costs, while substitutions between different token types are more heavily penalised. The algorithm is further extended to detect compound words, addressing errors from missing or inserted hyphens and spaces \cite{Kuhn2024}. During cost matrix computation, adjacent tokens can be merged to match tokens in the opposing text, enabling alignment of compounds with variable lengths and differing representations. While this dictionary-free approach is generally applicable, it only provides an approximation and ignores subtle differences in meaning. For instance, 'setup' (noun), 'set up' (verb) and 'set-up' (adjective) are not differentiated. Once the backtrace matrix is computed, the edit path is constructed as a sequence of operations linking reference and hypothesis tokens.

\subsection{Metrics}

The alignment results allow a quick calculation of the \ac{WER} by counting insertions, substitutions, and deletions while supporting complementary error rates covering punctuation and casing. Compound word errors can optionally contribute to the \ac{WER} score. \ac{NLP} classifications allow more granular evaluations, such as noun accuracy using \ac{POS} tags and proper name accuracy using \ac{NER} labels. Such attributes can isolate specific error types or apply weights in a combined evaluation score. Similarity measures such as Jaro-Winkler or cosine distance can be used to evaluate substitutions with finer precision \cite{Winkler1990, Roux2022}. Word-level information from an \ac{ASR} model allows for custom evaluations such as the relationship between confidence score and word correctness. All applied text normalisations are listed, and original values are preserved to support further analysis.

\subsection{Implementation}

OpenWER is developed in Python, the primary language in machine learning research, to facilitate community-driven improvements. Given the Levenshtein algorithm's $O(mn)$ complexity, several optimisations have been implemented to reduce processing time. Before computing the backtrace matrix, tokens are mapped to an optimised data structure, where each unique word is assigned an integer value for its original and lowercase forms, allowing for faster token comparison. Optional \ac{JIT} compilation further increases processing speed. The original tokens are restored when the alignment is constructed. Generic classes allow language-specific normalisations to be implemented with minimal programming knowledge, encouraging the open-source community to contribute their linguistic expertise. The repository includes datasets and evaluation scripts to systematically measure the impact of changes across different languages. All code and evaluation datasets are open-source.\footnote{https://github.com/shuffle-project/openwer} The repository includes examples of standard metrics and custom evaluations for straightforward adoption.

\subsection{Analysis}

We compare \ac{WER} results of our implementation with a popular library (JiWER\footnote{https://github.com/jitsi/jiwer}) to evaluate differences in accuracy measurement. To assess the impact of text normalisation, we evaluate both the JiWER normaliser, which follows a more conventional approach, and the Whisper normaliser, which applies more extensive transformations. We used the Common Voice 17 dataset for a broad cross-lingual comparison \cite{Ardila2020}. We transcribed the test split of all languages supported by two popular multilingual open-source \ac{ASR} models: Whisper (large-v3) and SeamlessM4T (v2-large) \cite{Radford2023, Barrault2023}. 52 languages were transcribed, totalling 691648 transcriptions, averaging 13300 samples per language ($SD = 11238$, $Min = 66$, $Max = 32804$).

We created a second dataset by transcribing 1000 randomly selected samples (seed=151) from the Common Voice 17 English test split with ten \ac{ASR} models. This dataset is used to validate OpenWER's ability to process \ac{ASR} word lists as input, and to provide an exemplary evaluation showing different metrics across models (see Table \ref{tab:metrics}). For each ASR system, the most accurate model was selected. Transcription jobs were performed in December 2024.

\section{Evaluation}

\begin{figure*}[t!]
  \centering
  \includegraphics[width=\linewidth]{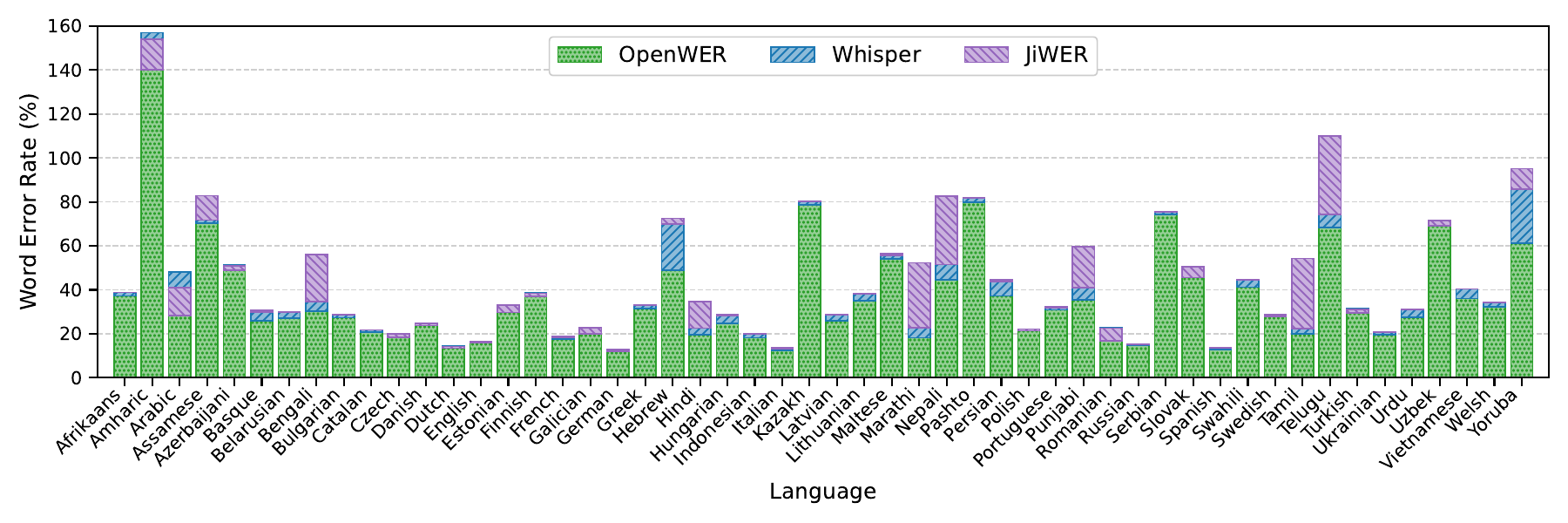}
  \caption{Word Error Rate comparison on Common Voice 17}
  \label{fig:results}
\end{figure*}

\subsection{Algorithm Robustness}

Compound word detection and punctuation tokens can alter the edit path in Levenshtein distance calculations, affecting \ac{WER} scores. To assess the robustness of OpenWER, we compare its results with those of the widely used JiWER library.

Without any text normalisation, case-sensitivity, and disabled compound word detection, the mean ($M$) \ac{WER} of OpenWER and JiWER was equal ($M=35.2\%$). A two one-sided test (TOST) was conducted to determine whether the effect of the Levenshtein algorithm on \ac{WER} was statistically equivalent to zero, given equivalence bounds of ±0.5\%. The test was significant, $t(691647) = 1.41$, $p < .001$, 90\% CI $[-0.2\%, 0.2\%]$, indicating that both implementations compute practically equivalent results for the standard Levenshtein distance. 

Applying common normalisation, where punctuation is removed, JiWER uses lowercasing, and OpenWER does not count casing errors, resulted in an average \ac{WER} of 30.1\% for OpenWER and 29.9\% for JiWER. The equivalence test with ±0.5\% bounds was significant, $t(691647) = 17.48$, $p < .001$, 90\% CI $[0.0\%, 0.3\%]$, indicating that both compute practically equivalent results for standard text normalisation.

To assess the effect of punctuation tokens on the edit path, the WER was measured using OpenWER with and without punctuation tokens (both $M=27.7\%$). The equivalence test with bounds of ±0.5\% was significant, $t(691647) = 23.53$, $p < .001$, 90\% CI $[-0.1\%, 0.2\%]$, indicating that the modified Levenshtein distance with variable editing costs per token type computes practically equivalent \ac{WER} scores while preserving punctuation tokens.

A paired samples t-test was conducted to compare \ac{WER} scores for OpenWER with enabled compound word detection ($M=25.3\%$) and without ($M=27.7\%$). The test showed a significant difference, $t(691647) = -156.61$, $p < .001$, $d = 0.04$. While the overall effect size suggests a small impact, the reduction in \ac{WER} varies considerably between languages. In some cases, the difference is minimal (0.1\%), but in others, it is up to 20\%. Results suggest that compound words are a major source of transcription errors in some languages but have a minor effect in others.

We evaluated whether different tokenisation methods produce equivalent WER scores using the second dataset containing transcription results from multiple \ac{ASR} models. We used OpenWER's basic tokeniser ($M=15.9\%$), spaCy as an \ac{NLP} tokeniser ($M=16.2\%$), and the \ac{ASR} word list output from each model ($M=15.9\%$). Pairwise equivalence tests with ±0.5\% bounds were significant between all methods: basic and \ac{NLP} ($t(9971) = -6.84$, $p < .001$, 90\% CI $[-1.4\%, 0.8\%]$); basic and word list ($t(9971) = -5.18$, $p < .001$, 90\% CI $[-1.1\%, 1.0\%]$); \ac{NLP} and word list ($t(9971) = -5.53$, $p < .001$, 90\% CI $[-1.3\%, 0.8\%]$).

We conclude that OpenWER produces equivalent \ac{WER} scores to JiWER when none or standard normalisation is applied. The modified Levenshtein algorithm can preserve punctuation tokens without affecting \ac{WER} and significantly reduce \ac{WER} by detecting compound words. Tokenisation methods can be used interchangeably if words are tokenised correctly.

\subsection{Word Error Rate}

Figure \ref{fig:results} shows the mean \ac{WER} per language, with each method using its full set of normalisations. OpenWER ($M=35.7\%$) consistently produced a lower or equal \ac{WER} compared to Whisper ($M=39.6\%$) and JiWER ($M=43.4\%$). The average \ac{WER} reduction of OpenWER compared to Whisper was 3.8\% (9.2\% relative) and compared to JiWER 7.7\% (14.5\% relative). The maximum \ac{WER} reduction of OpenWER compared to Whisper was 24.8\% (41.2\% relative) and compared to JiWER 41.4\% (65.1\% relative). A one-way ANOVA was conducted to determine the effect of the normalisation method on the resulting \ac{WER}. There was a significant effect in \ac{WER} between at least two methods, $F(2, 2074941 = 1214.961, p < .001)$. A Tukey's HSD test showed a significant difference between all methods (all $ps < .001$). These results highlight the effectiveness of OpenWER in reducing \ac{WER} across languages and demonstrate its advantages over existing normalisation methods, particularly for some low-resource languages.

\subsection{Exemplary Evaluation}

\begin{table}[t!]
    \centering
    \caption{Accuracy of ASR models (*** indicates $p < .001$)}
    \begin{tabular}{lrrrc}
        \toprule
        Model & WER$\downarrow$ & Punct.$\downarrow$ & Casing$\downarrow$ & $r_{pb}\uparrow$ \\
        \midrule
        AWS & 11.1 & 11.5 & 5.4 & 0.148*** \\
        AssemblyAI & 10.3 & 17.7 & 9.4 & 0.362*** \\
        Deepgram & 20.3 & 24.1 & 8.8 & 0.485*** \\
        Google & 11.1 & 15.5 & 7.9 & 0.278*** \\
        IBM & 22.0 & 98.3 & 83.5 & 0.107*** \\
        Microsoft & 12.7 & 15.0 & 6.7 & 0.179*** \\
        Rev AI & 24.7 & 29.0 & 14.9 & 0.459*** \\
        SeamlessM4T & 25.1 & 13.0 & 4.6 & - \\
        Speechmatics & 8.8 & 21.1 & 9.8 & 0.446*** \\
        Whisper & 12.9 & 21.6 & 9.2 & 0.517*** \\
        \bottomrule
    \end{tabular}
    \label{tab:metrics}
\end{table}

Table \ref{tab:metrics} shows the evaluation results for several \ac{ASR} models using OpenWER. The analysis is based on a small dataset and is intended to illustrate the capabilities of OpenWER rather than providing a definitive comparison of ASR models. The results show that models with a similar \ac{WER} can differ notably in terms of punctuation and casing accuracy. This highlights the importance of conducting comprehensive assessments of transcription accuracy. As an exemplary use case of metadata embedding, we computed the point-biserial correlation ($r_{pb}$) between word correctness and confidence score. SeamlessM4T does not provide word-level confidence scores and was excluded. The results show a moderate correlation at best, suggesting that confidence scores from these \ac{ASR} models were not sufficiently reliable for determining word correctness.

\subsection{Performance}

JiWER leverages a C implementation of the Levenshtein distance, whereas OpenWER is implemented purely in Python. To compare performance, we computed Levenshtein distances for token counts from 1 to 1000 in steps of 25 and averaged the processing time for each length over 25 iterations. JiWER processed $4618.9$ tokens/ms and OpenWER $1.9$ tokens/ms. \ac{JIT} compilation increased the processing speed of OpenWER to $34.8$ tokens/ms. Still, JiWER remains superior in speed, highlighting the limitations of OpenWER for performance-critical use and the potential benefits of implementing the modified Levenshtein algorithm in a compiled language.

\section{Conclusion}

Objective metrics are essential to measure the transcription accuracy of \ac{ASR} models across languages and to identify performance gaps, particularly in low-resource languages. \ac{WER} has been criticised for its limitations, but alternative metrics have yet to see wider adoption. OpenWER addresses these challenges and improves \ac{WER}'s reliability by incorporating language-specific normalisation and compound word detection. Our evaluation demonstrated significant accuracy gains across multiple languages. The flexible, token-based approach preserves complementary metrics for punctuation and casing, and allows the use of metadata for novel evaluations. Through open-source and its modular design, OpenWER aims to contribute to fairness in multilingual evaluation, encourage community-driven improvements, and provide a foundation for future accuracy metrics that address the shortcomings of \ac{WER}.

\section{Acknowledgements}

This work was conducted as part of the SHUFFLE Project and funded by “Stiftung Innovation in der Hochschullehre”.

\bibliographystyle{IEEEtran}
\bibliography{mybib}

\end{document}